\documentclass[journal]{IEEEtran}
\usepackage{multicol}
\usepackage{color}
\usepackage{xspace}
\usepackage{times}
\usepackage{epsfig}
\usepackage{graphicx}
\usepackage{amssymb}
\usepackage{amsmath}
\usepackage{algorithm}
\usepackage{algorithmic}

\makeatletter
\g@addto@macro\normalsize{
	\setlength\abovedisplayskip{1pt}
	\setlength\belowdisplayskip{0pt}
	\setlength\abovedisplayshortskip{0pt}
	\setlength\belowdisplayshortskip{0pt}
}
\makeatother

\ifCLASSINFOpdf
\else
\fi
\begin{document}
%
\title{Exemplar-AMMs: Recognizing Crowd Movements from Pedestrian Trajectories}
%
%
%

\author{Wenxi~Liu,
		~Rynson W.H. Lau,
		~Xiaogang Wang,
		~Dinesh Manocha

}

%
%

\markboth{Journal of \LaTeX\ Class Files,~Vol.~13, No.~9, September~2014}%
{Shell \MakeLowercase{\textit{et al.}}: Bare Demo of IEEEtran.cls for Journals}
%




\maketitle

\begin{abstract}
In this paper, we present a novel method to recognize the types of crowd movement from crowd trajectories using agent-based motion models (AMMs). Our idea is to apply a number of AMMs, referred to as exemplar-AMMs, to describe the crowd movement. Specifically, we propose an optimization framework that filters out the unknown noise in the crowd trajectories and measures their similarity to the exemplar-AMMs to produce a crowd motion feature.  We then address our real-world crowd movement recognition problem as a multi-label classification problem. Our experiments show that the proposed feature outperforms the state-of-the-art methods in recognizing both simulated and real-world crowd movements from their trajectories. Finally, we have created a synthetic dataset, \textit{SynCrowd}, which contains 2D crowd trajectories in various scenarios, generated by various crowd simulators. This dataset can serve as a training set or benchmark for crowd analysis work.
%
\end{abstract}


%
\IEEEpeerreviewmaketitle

\section{Introduction}

With more cameras available everywhere in recent years, a large number of videos are captured, not only for entertainment but also for surveillance.
As people are often the main subject of interest in these videos and they usually show up in groups, many researchers are interested in understanding the collective behaviors of groups of people, and studying crowd behaviors for video-based applications like social event/action recognition~\cite{Wu2014TMM,Yang2015TMM}, learning motion features for pedestrian tracking~\cite{pellegrini_youll_2009,Yuan2015TMM}, and retrieval in surveillance datasets~\cite{Feris2012TMM,Presti2012TMM}.


A fundamental problem in crowd behavior analysis is recognizing types of crowd movements.
Crowd movement recognition has been the subject of computer vision based works because it is critical for understanding crowd behaviors and identifying social events in video surveillance.
However, this is a challenging research topic, because crowd movement patterns are complex and affected by many factors, including crowd density, scene configuration, and crowd psychology.

In this paper, we propose an approach to recognize crowd movements based on crowd trajectories. In general, each crowd trajectory represents the spatial-temporal information of one pedestrian in a crowd. Crowd trajectories are informative for analyzing the mutual interactions among pedestrians in a crowd, e.g., how actively pedestrians react to oncoming pedestrians. Leveraging such latent information can effectively improve the accuracy of crowd recognition.
The difficulty lies in how to capture crowd trajectories. In recent years, pedestrian tracking techniques have made significant advances. As a result, capturing crowd trajectories with minor manual inference is now possible.
Previous works on trajectories-based crowd analysis mostly concern about trajectory clustering~\cite{ge2012group}, semantic region inference~\cite{wang2011trajectory,zhou2012understanding}, or retrieval in trajectory datasets~\cite{Presti2012TMM}. To the best of our knowledge, none of the existing work uses trajectories for crowd movement recognition.


\begin{figure*}
	\centering
	\includegraphics[width=0.8\textwidth]{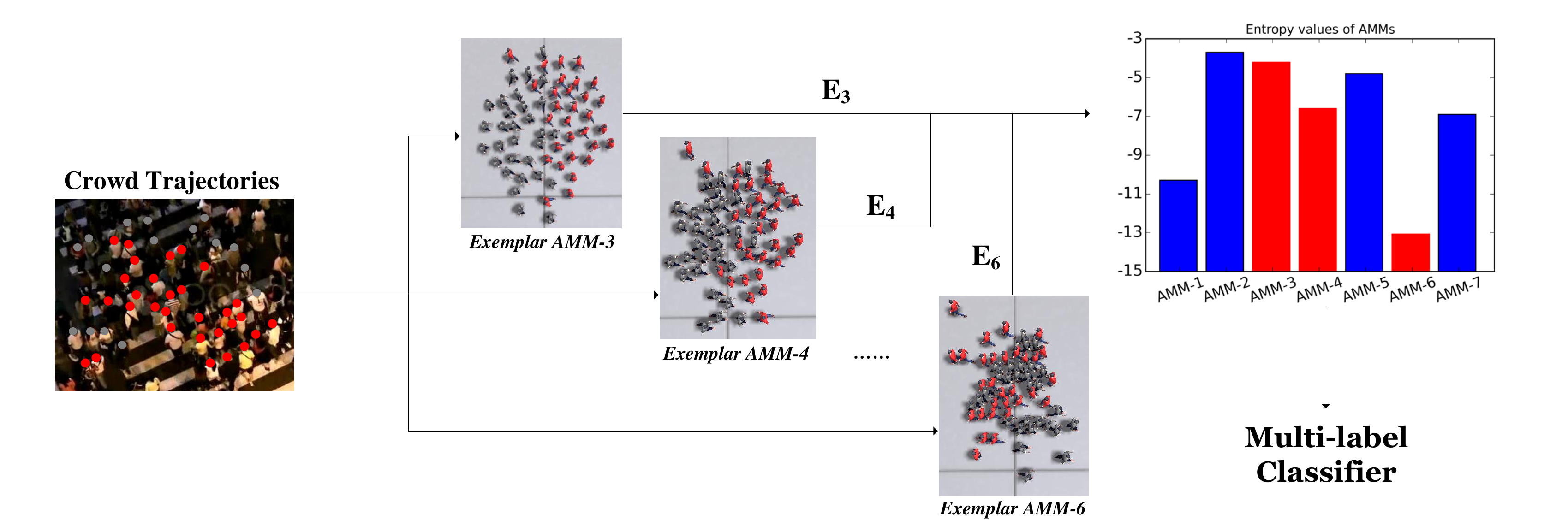}
	\caption{Procedure for computing the proposed crowd motion feature. Crowd trajectories (red and gray dots indicating the tracked pedestrians) and multiple exemplar-AMMs, which are capable of simulating different crowd behaviors, are given. Here, we show the screenshots of rendered simulated crowd movements from three different AMMs. We can see that the agents in the simulated crowd movement of \textit{Exemplar AMM-3} are repulsive (i.e., each agent maintains a secure distance from others). Agents of \textit{Exemplar AMM-4} are less repulsive, but still produce a clogging situation, while agents of \textit{Exemplar AMM-6} are more aggressive (i.e., not afraid of collisions) in moving towards their own destinations. Our feature is computed as the similarity between the real-world crowd trajectories with those of each AMM, denoted by the entropy values (e.g., $\textbf{E}_3$, $\textbf{E}_4$, and $\textbf{E}_6$). The lower the entropy value, the more similar they look (i.e., the input crowd movement is closer to that of \textit{Exemplar AMM-6}). Hence, these entropy values can jointly classify the type of the input crowd movement.}
	\label{fig:framework}
\end{figure*}

To recognize crowd trajectories, we prefer a feature that takes into account not only the global motion pattern but also the latent interaction attributes.
Further, we need to refine the crowd trajectories at runtime, given that
they are compounded with unknown noise (e.g., measurement errors and errors due to the approximated perspective transformation from the image-space to the ground-space).
To address these problems in this work, we leverage~\textit{Agent-based Motion Models} (AMMs) \cite{reynolds_flocks_1987,helbing_social_1995,van_den_berg_reciprocal_2011}, which have recently been shown to be effective at modeling the interactions in crowds.
We measure how well a certain AMM can simulate the crowd data. Because each AMM is able to model a specific interaction behavior of crowds, if the AMM can fit the input crowd data well, the crowd data probably contains the interaction behavior that the AMM models.
Meanwhile, we can also filter out the unknown noise of the input crowd trajectories with the assistance of the AMM.
Specifically, we extend the algorithm in~\cite{guy2012statistical} (which is used to quantify how well crowd simulators perform in synthesizing virtual scenes) to measure the similarity between any given AMM and the input crowd trajectories. However, we remove their assumption of homogeneity, since crowds usually consist of different types of individuals. This allows us to investigate the latent mutual interaction for each crowd member.

In addition, as a single AMM cannot model crowd behaviors in an omnipotent way, leveraging a single AMM to recognize different types of crowd movements is not robust.
On the other hand, training a specific AMM for each crowd scenario is also inefficient.
To obviate this difficulty in selecting or training an AMM, we are inspired by
prior works on object recognition (e.g., \cite{malisiewicz2011ensemble}) that leverage exemplar models to infer unknown models.
We introduce a similar exemplar-based framework that uses multiple distinct AMMs (or the same model with different parameters) as exemplar models to jointly measure crowd trajectories. The similarity measurements from these exemplar-AMMs serve as the crowd motion feature.


To evaluate our proposed feature, crowd movement recognition is formulated as a multi-class classification problem. However, crowd movement may contain several attributes at the same time. For example, coherent crowd behavior may be blended with group swapping motion, which is often observed in crosswalks. Hence, we treat the crowd movement recognition problem in this paper as a multi-label classification problem; i.e., each instance can own multiple class labels.

\textbf{Main Results}: In this paper, we propose a novel method to leverage multiple exemplar-AMMs for crowd movement recognition based on pedestrian trajectories. In particular, we present a framework that can measure crowd movements numerically by filtering out the noise of the trajectories, as shown in Fig.~\ref{fig:framework}. The proposed algorithm can produce an entropy descriptor that evaluates crowd movement with reference to any given AMM. All of the entropy descriptors from the exemplar-AMMs are combined to form a robust middle-level feature of the crowd movement. We evaluate this feature by performing multi-label classification experiments in both simulated crowd trajectories and real-world crowd movement. To study the feature, we further produce a synthetic crowd dataset, \textit{SynCrowd}, consisting of various types of simulated crowd movements, which can be used as the training dataset or the benchmark for further crowd analysis research.


\section{Related Works}

In this section, we summarize prior works on visual analysis of crowds and on agent-based motion models.

\vspace{-.13in}
\subsection{Crowd Analysis}

Prior works on crowd analysis can generally be categorized into holistic methods, particle-based or feature point-based methods, and individual-based methods.

\noindent\textbf{Holistic methods:} These methods treat a crowd as an aggregated whole \cite{chan2008modeling, mahadevan2010anomaly, wang2009unsupervised, kuettel2010s}.
Chan \emph{et al.} \cite{chan2008modeling} present a method based on dynamic textures. It represents video sequences as observations from a linear dynamical system. Mahadevan \emph{et al.} \cite{mahadevan2010anomaly} extend the idea to anomaly detection in crowd motion.
There are also many works that handle low-level visual features, e.g., optical flow or moving pixels, and build up topic models to discover various aggregated motion patterns in crowded scenes \cite{wang2009unsupervised, kuettel2010s}.
These methods can effectively analyze global visual features, but their models are not generalizable to different scenarios because they need to be retrained for different crowd movements.
Another drawback of these methods is that most of the information regarding individuals is not available.

\noindent\textbf{Individual-based methods:} These methods treat a crowd as
a collection of individuals, rather than as global patterns, and consider their interactions.
They mainly analyze trajectories and use the complete or partial spatial-temporal information of individuals for analysis \cite{wang2011trajectory,zhou2012understanding,choi2012unified,li2013recognizing}. Wang \emph{et al.} \cite{wang2011trajectory} present an approach for unsupervised trajectory analysis and semantic region modeling. Zhou \emph{et al.} \cite{zhou2012understanding} propose a mixture model to learn motion patterns and predict pedestrians' behaviors from the partially-observed trajectories. Choi \emph{et al.} \cite{choi2012unified} set up a hierarchical activity model to recognize collective activities and the interaction between targets based on the individual trajectories.
These methods investigate motion patterns or individuals' interactions accurately, but the learned models are not
generalizable to different scenes either. To model crowd motion, some works also study real crowd trajectories \cite{wolinski2014parameter,charalambous2014data}. Wolinski \emph{et al.} \cite{wolinski2014parameter} apply genetic algorithm to fit crowd data. Charalambous \emph{et al.} \cite{charalambous2014data} propose a rankable metric to measure individual trajectories.
Compared with prior trajectory-based work, we leverage AMMs to extract middle-level motion features for real crowd movement recognition.

\noindent\textbf{Particle-based or feature point-based methods:} These methods analyze high-density crowd movements~\cite{ali_floor_2008,ali2007lagrangian,mehran2009abnormal,ali2013measuring}.
Ali \cite{ali2013measuring} presents an approach based on a particle-based representation to explicitly take into consideration the interactions among objects while measuring the flow complexity.
Like particle-based methods, the feature point-based methods deal with tracks generated by trackers \cite{zhou2013measuring, shao2014scene}.
Zhou \emph{et al.} \cite{zhou2013measuring} analyze the collective crowd movements by measuring path similarities among crowds on the collective manifold.
Shao \emph{et al.} \cite{shao2014scene} present a method for detecting groups from the tracks to analyze the fundamental group properties, which can be easily generalized to different crowd systems. The resulting grouping profile of a crowd is used for analysis.
These methods provide a trade-off between holistic methods and individual-based methods. Their limitation is that
individual information cannot be fully discovered, as the trajectories captured by feature-point trackers are usually segmented and affected by the background noise or object poses.

\vspace{-.12in}
\subsection{Agent-based Motion Models (AMMs)}

AMMs, which primarily model local collision avoidance or local behaviors of pedestrians in crowds, are usually used as predictors or simulators.
Many AMMs have been proposed, including local rule-based models \cite{reynolds_flocks_1987,reynolds_steering_1999,Moussaid2011}, the Social Force model \cite{helbing_social_1995}, and geometry-based algorithms \cite{ondej_synthetic-vision_2010,pettre_experiment-based_2009,Karamouzas2010,van_den_berg_reciprocal_2011}.
Generally speaking, AMMs focus on the spatial location, rather than the gesture or posture, of each crowd member (or agent), and most of them simplify agents to circles in 2D space.
AMMs usually are controlled by several motion parameters, $\beta$,
e.g., the number of the nearby persons and the maximum speed.
Hence, any AMM can be formulated as a non-linear function $f$ tha estimates the crowd state at the next timestep from the current crowd state $\textbf{x}_t$, i.e., $\textbf{x}_{t+1} = f_{\beta}(\textbf{x}_t)$.

\begin{figure}
	\centering
	\begin{tabular}{c@{}c@{}c@{}c}
		{\footnotesize a)}
		\includegraphics[width=0.25\linewidth]{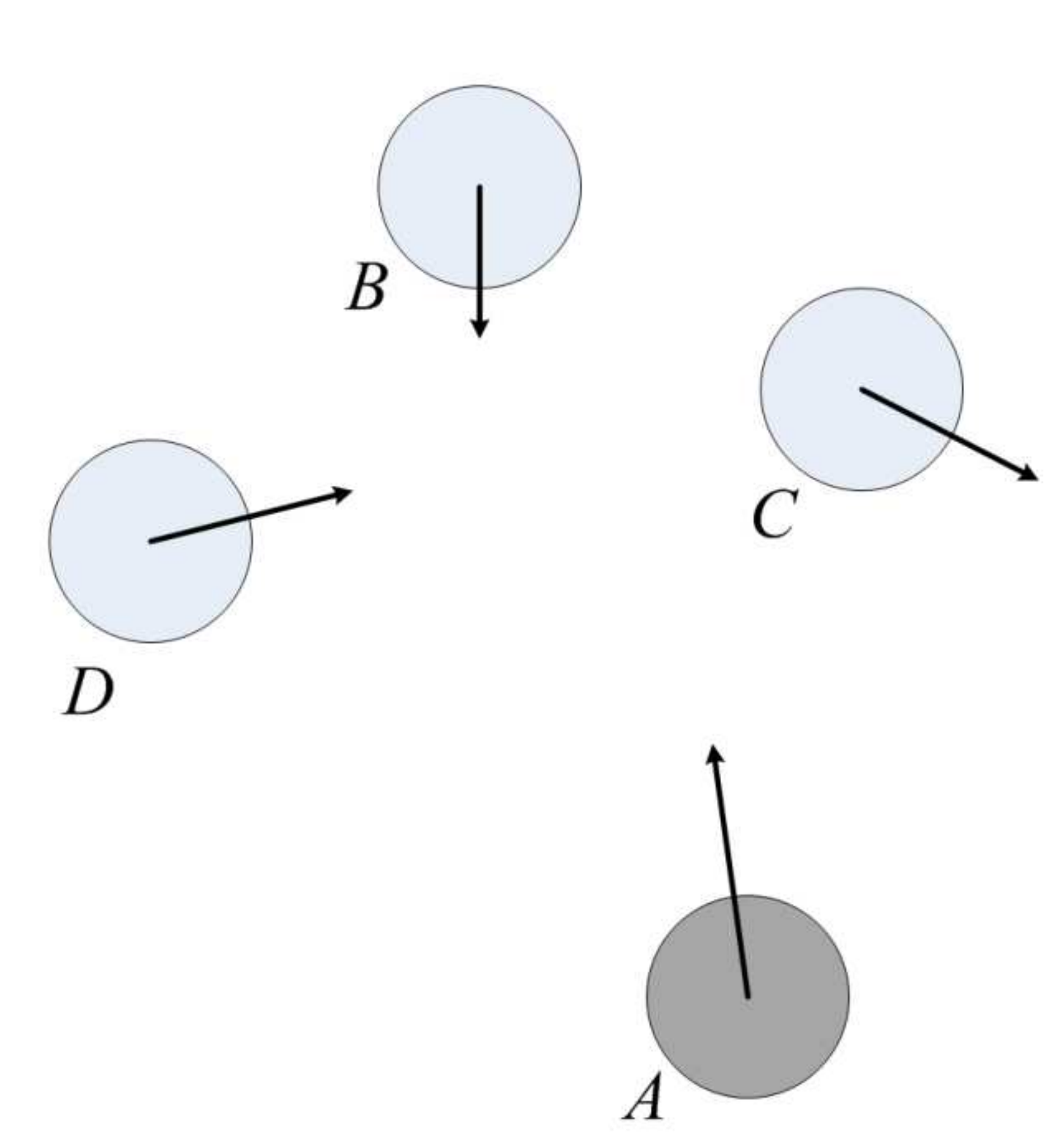} &
		\hspace{.5in}{\footnotesize b)}
		\includegraphics[width=0.25\linewidth]{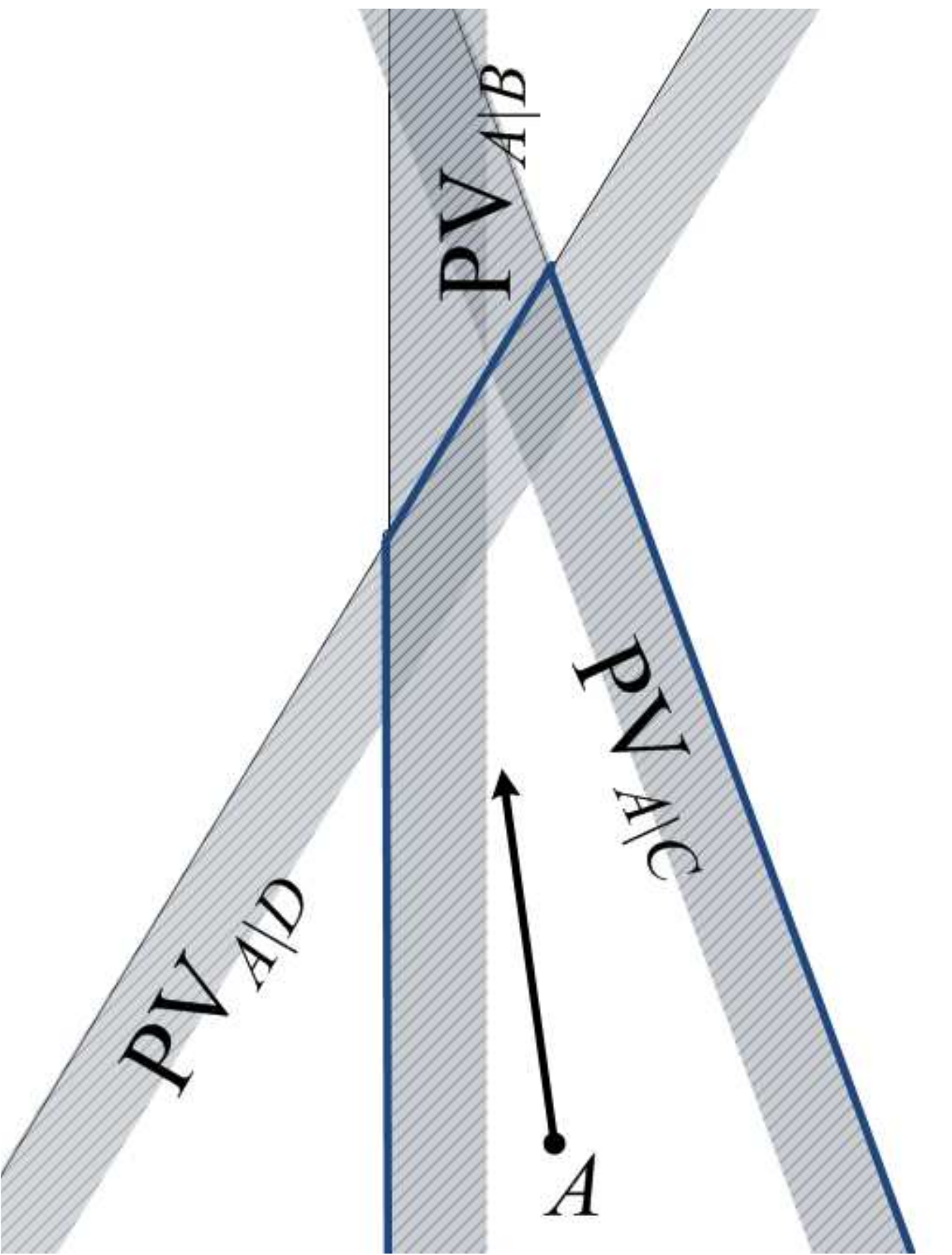}
	\end{tabular}
	\caption{The RVO multi-agent simulation. (a) shows four agents in a 2D space, with arrows indicating their velocities. (b) shows the corresponding velocity space. The shaded side of each plane is the set of permitted velocities for agent $A$ to avoid collision with the other corresponding agent. The region with bolded lines denotes the collision-free velocity set of agent $A$.}
	\label{fig:rvo}
	\vspace{-.13in}
\end{figure}

\textit{Reciprocal Velocity Obstacle (RVO)}:
As one of many geometry-based methods, the RVO model~\cite{van_den_berg_reciprocal_2011} provides real-time collision-free crowd simulation, which is suitable for our work.
It predicts agents' positions and velocities in a 2D ground space, given the states of other agents at the current timestep, and makes sure that they will not lead to collision among the agents. In Fig.~\ref{fig:rvo}, agent $A$ chooses the optimal velocity from the permitted velocity set to avoid collision. Guy \emph{et al.} \cite{guy2011simulating} use RVO to model different crowd behaviors by adjusting its parameters.


\textit{Data-driven AMMs}: In order to learn crowd models, many data-driven AMMs have been proposed in recent years \cite{lee2007group,charalambous2014pag,bera2015gi}.
Unlike these data-driven models, our framework utilizes multiple exemplar-AMMs, which avoids training AMMs.

\textit{AMMs in Computer Vision}: AMMs are mainly used for multi-target tracking \cite{pellegrini_youll_2009,yamaguchi_who_2011,wenxiliu2015tracking} and anomaly detection \cite{mehran2009abnormal} in computer vision. In tracking, AMMs are usually integrated with some appearance models to predict pedestrians' future positions.
Antonini \emph{et al.} \cite{antonini_behavioral_2006} present a Discrete Choice Model-based motion model to discretize the velocity space of a pedestrian and to model the choosing of the optimal velocity for the pedestrian,
while Pellegrini \emph{et al.} \cite{pellegrini_youll_2009} formulate pedestrians' movements as an energy optimization problem that factors in navigation and collision avoidance.
Yamaguchi \emph{et al.} \cite{yamaguchi_who_2011} apply
insights from the social force approach to estimate future positions of the pedestrians.
In anomaly detection, Mehran \emph{et al.} \cite{mehran2009abnormal} apply the social force model to represent the abnormal motion of the crowd.
In all of these works, the efficacy of the motion model depends on the selection of the motion parameters. To overcome this problem, we propose a multiple-exemplar-model framework, which uses several representative AMMs to jointly analyze the input crowd data.

\section{Exemplar-AMMs based Crowd Feature}

Similarly to previous exemplar-based works, our method needs to measure the connection between crowd-motion data and each exemplar model.
However, as crowds are complex by nature, it is difficult to directly compare crowd motion data
with other crowd motion examples, even if the individual trajectories are provided.
%
Our observation is that AMMs can model (simulate or predict) real crowd movements, but usually
perform differently for varied types of crowd interactions. For example,
some AMMs are good at modeling collective behavior~\cite{vicsek1995novel} and others at collision avoidance~\cite{helbing_social_1995,van_den_berg_reciprocal_2011}. Some can simulate only aggressive behavior and others conservative behavior~\cite{guy2011simulating}.
%
Hence, different AMMs can be used to describe different aspects of crowd movement.
In this paper, we apply such observations in the following way. For each exemplar-AMM, an entropy descriptor is computed to measure its similarity to the given
crowd trajectories. All entropy descriptors computed from multiple exemplar-AMMs are then grouped as a robust crowd-motion feature. Our method is illustrated in Fig.~\ref{fig:framework}.

In the following subsections,
we first explain generally how we model AMM and the crowd data, and then introduce how we compute the entropy descriptors.

\subsection{AMMs and Crowd Data}
\label{subsec:model}


\begin{table}
	\centering
	\caption{The definitions of the notations used.}
	\begin{tabular}{|c|l|}
		\hline
		\textbf{Symbol} & {\centering \textbf{Definition}} \\
		\hline\hline
		$N$ & Number of agents \\
        \hline
		$T$ & Number of frames of the crowd movement\\
		\hline
		$\textbf{p}_i,\textbf{v}_i,\textbf{d}_i$ &
        The position, velocity and desired velocity of $i^{th}$ agent, \\
        & where $\textbf{p}_i,\textbf{v}_i,\textbf{d}_i\in \mathbb{R}^{2}$\\
		\hline
		$\textbf{x}_t$ & Crowd state vector, i.e.,\\
		&  $\textbf{x}_t = [\textbf{p}_1,\textbf{v}_1,\textbf{d}_1,...,\textbf{p}_N,\textbf{v}_N,\textbf{d}_N] \in \mathbb{R}^{6N}$\\
		\hline
		$\textbf{z}_t$ & Observed crowd state (i.e. pedestrians' positions),\\
		&  $\textbf{z}_t = [\textbf{p}_1,...,\textbf{p}_N] \in \mathbb{R}^{2N}$ \\
		\hline
		$\mathcal{X}$, $\mathcal{Q}$, $\mathcal{R}$ & Gaussian distributions\\
        \hline
		$\textbf{q}$, $\textbf{r}$ & Random samples drawn from $\mathcal{Q}$ and $\mathcal{R}$, respectively \\
		\hline
	\end{tabular}
	\label{tab:terms}
\end{table}

For a given AMM, the state $\textbf{x}_t$ of the crowd at a specific timestep $t$ contains the status of all agents: positions, velocities, and desired velocities. (See Table~\ref{tab:terms} for a summary of the major notations).
An AMM can be treated as a non-linear function $f$ that propagates $\textbf{x}_t$ forward to the next timestep,
$\textbf{x}_{t+1} = f(\textbf{x}_t)$.
However, it cannot always make exact accurate predictions.
We denote the true crowd state as $\dot{\textbf{x}}_{t}$ (usually unknown) and the error of the prediction as $\textbf{r}_t$:
\begin{align}
\label{eq:start}
\dot{\textbf{x}}_{t+1} = f(\textbf{x}_t) + \textbf{r}_t.
\end{align}
The crowd data contains all agents' temporal-spatial positions (or trajectories), represented as $T$ vectors in $\mathbb{R}^{2N}$, $\{\textbf{z}_1,\dots \textbf{z}_T\}$, i.e., observations. Since they are often compounded with noise, they may be considered as a noisy projection of the true crowd state $\{\dot{\textbf{x}}_1, \dots \dot{\textbf{x}}_T\}$ at timestep:
$\textbf{z}_t = h(\dot{\textbf{x}}_t) +\textbf{q}_t$ ($\textbf{q}_t \sim \mathcal{Q}$),
where $h$ projects crowd state $\dot{\textbf{x}}_{t} \in \mathbb{R}^{6N}$ to $\textbf{z}_t \in \mathbb{R}^{2N}$, keeping only the position information. $\textbf{q}_t$ represents the observation noise, subject to a zero-mean Gaussian distribution. We assume $\mathcal{Q}$ to be static and measurable. In practice, we usually assign it a small scale.

\vspace{-.1in}
\subsection{Entropy Descriptor}

To introduce our descriptor, we first assume that the prediction error $\textbf{r}_t$ in Eq.~\ref{eq:start} is subject to a zero-mean Gaussian distribution $\mathcal{R} = \mathcal{N}(\textbf{0}, \Sigma)$. By measuring the scale of $\mathcal{R}$, we can quantify the similarity between the given AMM and the data. A larger scale implies a larger divergence between the AMM and the data, and vice versa.

As mentioned above, given the noisy observations $\textbf{z}_t$ ($t \in \{1,\dots T\}$) containing pedestrians' observed positions only, the true crowd state $\dot{\textbf{x}}_t$ (which contains all pedestrians' positions, velocities, and desired velocities) is often uncertain.
To account for this uncertainty, we represent the crowd state as a Gaussian distribution: $\mathcal{X}_t = \mathcal{N}(\dot{\textbf{x}}_t, \cdot)$ ($t \in \{1, \dots T\}$).

With regard to the prediction error $\Sigma$ and the crowd state distribution $\mathcal{X}_t$, our goal is to maximize the log-likelihood of $\mathcal{R}$ with a given AMM $f$, i.e.
\begin{align}
\label{eq:obj}
\mathcal{X}_t,\Sigma &= \arg\max_{\mathcal{X}_t,\Sigma} \ell \ell (\mathcal{R} ).
\end{align}

\vspace{-.1in}
\subsection{Optimization}

The objective function of Eq.~\ref{eq:obj} has decisive variables: the prediction error $\Sigma$ and the crowd state distribution $\mathcal{X}_t$.
We adopt an optimization strategy similar to the Expectation-Maximization algorithm. We first
optimize $\mathcal{X}_t$ based on observations and then maximize the likelihood of $\Sigma$.
Thus, the following two steps are performed iteratively:

\textbf{STEP-1: Fix $\Sigma$ ($\mathcal{R}$ is known) and optimize $\mathcal{X}_t$.}
Since the prediction error $\mathcal{R}$ is known and the noisy observation $\textbf{z}_t$ ($t \in \{1 \dots T\}$) is given, we aim to compute the optimal crowd states, i.e. removing the observation noise of crowd trajectories and estimating the states (i.e., positions, velocities, and desired velocities) of pedestrians.

As mentioned, we assume the crowd states as distributions, $\mathcal{X}_t$ ($t \in \{1 \dots T\}$). We then apply the Ensemble Kalman Filter (EnKF) algorithm~\cite{evensen2003ensemble} to compute the optimal values of crowd state distributions. In EnKF, $\mathcal{X}_t$ is represented by a set of samples, or ensembles,
i.e., $[\textbf{x}_t^1, \dots, \textbf{x}_t^M] \sim \mathcal{X}_t$, where $M$ is the number of the ensembles.

Similarly to the Kalman filter, in each timestep $t$ ($t \in \{1 \dots T \}$), EnKF consists of two main steps: 1) \textit{prediction} and 2) \textit{correction}. In \textit{prediction}, given the current crowd state at $t$, we leverage the provided AMM, $f$, and the fixed prediction error, $\mathcal{R}=\mathcal{N}(\textbf{0}, \Sigma)$, to sequentially predict the following crowd states. Specifically, the next crowd state distribution is predicted subject to $f$, i.e. $\hat{\mathcal{X}}_{t+1} = \mathcal{N}(f(\mathcal{X}_t), \Sigma)$. Since the crowd state distribution is represented by ensembles, each of them evolves via the AMM in addition to random Gaussian noise drawn from $\mathcal{R}$, i.e., $[f(\textbf{x}_t^1)+\textbf{r}_t^1, \cdots, f(\textbf{x}_t^M)+\textbf{r}_t^M]$, where $\textbf{r}_t^m \sim \mathcal{R}$ ($m \in \{1 \dots M\}$).
In \textit{correction}, the evolved crowd state distribution $\hat{\mathcal{X}}_{t+1}$ is corrected by the observations, $\textbf{z}_{t+1}$ (i.e. positions of crowd members).
According to~\cite{evensen2003ensemble}, the posterior ensemble is similar to the Kalman filter:
\begin{align}
\label{eq:enks1}
\mathcal{X}_{t+1} = \hat{\mathcal{X}}_{t+1} + \textbf{K}(\textbf{z}_{t+1} - h(\hat{\mathcal{X}}_{t+1})),
\end{align}
\noindent where $h$ is the projection function and $\textbf{K}$ is the Kalman gain matrix.
To compute the Kalman gain matrix, we define $A = \hat{\mathcal{X}}_{t+1} - \frac{1}{M}\sum \hat{\mathcal{X}}_{t+1}$ and $B = h(\hat{\mathcal{X}}_{t+1}) - \frac{1}{M}\sum h(\hat{\mathcal{X}}_{t+1})$.
The Kalman gain matrix can be computed as $\textbf{K} = (M-1)^{-1} A B^T P^{-1}$, where $P = (M-1)^{-1} B B^T +\Sigma_Q$ and $\Sigma_Q$ is the known covariance matrix of $\mathcal{Q}$. Integrating $\textbf{K}$ into Eq.~\ref{eq:enks1}, the crowd state distribution at the next time step is updated. Hence, by sequentially predicting and correcting crowd states, the noise of the trajectories are removed as much as possible and we approximately obtain the optimal crowd states.

\textbf{STEP-2: Fix $\mathcal{X}_t$ and optimize $\Sigma$.}
As mentioned, the prediction error $\Sigma$ is the covariance of zero-mean distribution $\mathcal{R}$.
Here we assume that given an AMM, the prediction error for each individual is independent. Therefore, the distribution of the AMM's prediction error $\mathcal{R}$ is formulated as:
\begin{align}
\label{eq:r}
\mathcal{R} & = \mathcal{N}(\textbf{0}, \Sigma) \nonumber \\
&= \mathcal{N}(\textbf{0},
\begin{bmatrix}
\Sigma_1 &   &  \\
  & \ddots &   \\
  &  & \Sigma_N
\end{bmatrix}),
\end{align}

\noindent where $\Sigma_i \in \mathbb{R}_{6 \times 6}  \text{ } (i \in \{1,...,N\})$ represents the individual covariance of each agent in the crowd.
To find the optimal $\Sigma$, we maximize the expected log-likelihood of $\Sigma$.
Its maximum likelihood estimation can be computed as:
\begin{align}
\label{eq:mle}
\hat{\Sigma} &= \frac{1}{T-1}\sum_{t=1}^{T-1} (\mathcal{X}_{t+1} - f(\mathcal{X}_t))(\mathcal{X}_{t+1} - f(\mathcal{X}_t))^T\\
&= \frac{1}{(T-1)M}\sum_{t=1}^{T-1}\sum_{m=1}^{M} (\textbf{x}_{t+1}^m - f(\textbf{x}_{t}^m))(\textbf{x}_{t+1}^m - f(\textbf{x}_{t}^m))^T.\nonumber
\end{align}

%

\begin{figure*}
	{\centering
		\begin{tabular}{c@{}c@{}c@{}c@{}c@{}c}
			\includegraphics[width=.14\textwidth]{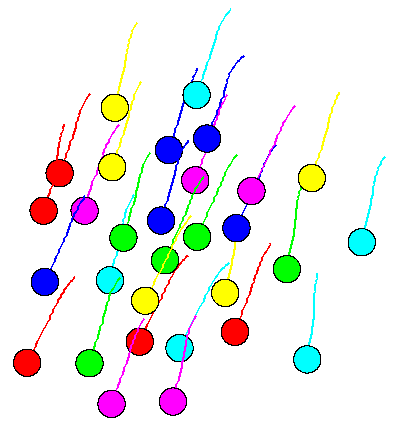} &		
			\includegraphics[width=.14\textwidth]{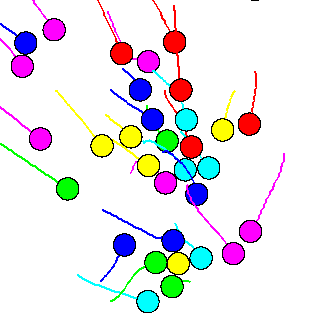} &
			\includegraphics[width=.14\textwidth]{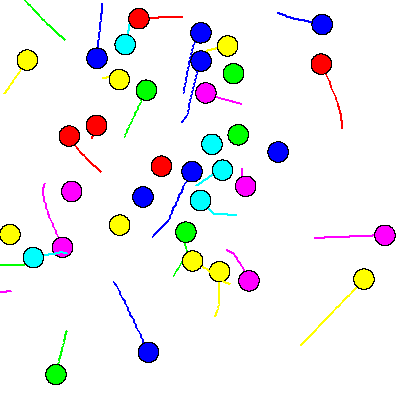} &
			\includegraphics[width=.14\textwidth]{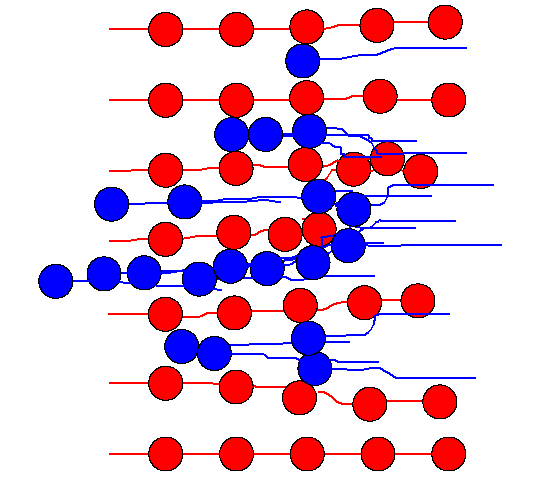} &
			\includegraphics[width=.14\textwidth]{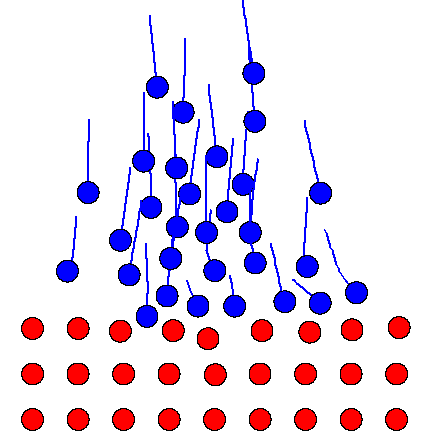} &
			\includegraphics[width=.25\textwidth]{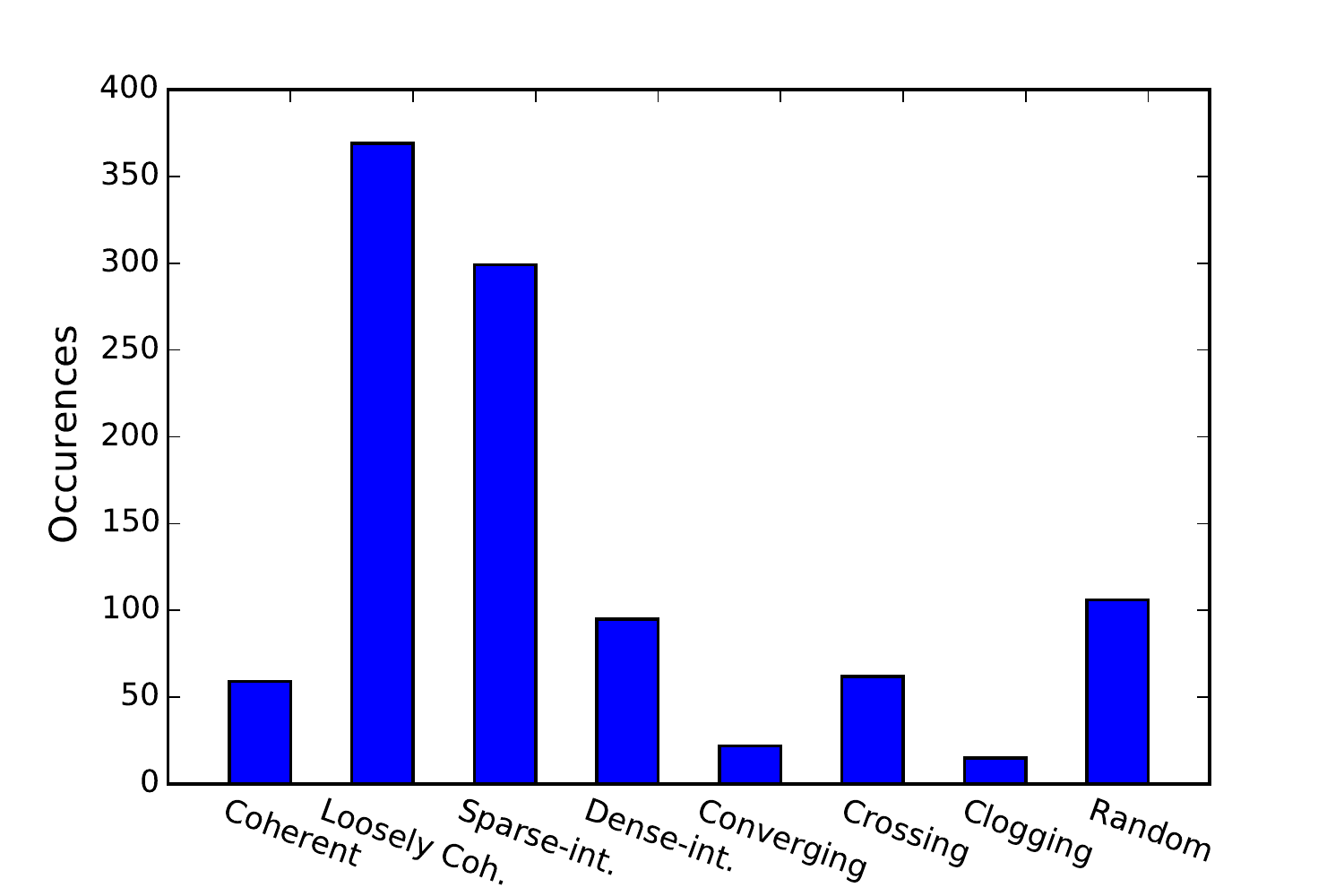}\\
			{\footnotesize (a) Coherent motion} & {\footnotesize (b) Converging motion} & {\footnotesize (c) Sparse interaction} & {\footnotesize (d) Crossing} & {\footnotesize (e) Clogging} & {\footnotesize (f) Scene-category statistics of real-world dataset}
		\end{tabular}
		\caption{(a-e) Examples of 2D simulation crowd movements from \textit{SynCrowd}. (f) The statistics of the crowd movement in the real-world dataset.}
		\label{fig:examples_motions}
	}
	\vspace{-.13in}
\end{figure*}

\vspace{-.13in}
\subsection{Entropy Computation}

By performing STEP-1 and STEP-2 iteratively, the algorithm will converge, since both steps optimize the objective function explained in Eq.~\ref{eq:obj}.
Thus, we can estimate the optimal crowd state $\mathcal{X}_t$ and the prediction error distribution $\mathcal{R}$. Consequently, the estimated individual covariance matrices, $\hat{\Sigma}_i \text{ } (i \in \{1,...,N\})$, are also computed. Hence, the individual entropy values of the crowd are computed as $\textbf{E} = [\frac{1}{2}\ln |(2 \pi e) \hat{\Sigma}_1|, \dots, \frac{1}{2}\ln |(2 \pi e) \hat{\Sigma}_N|]^T \in \mathbb{R}^{N \times 1}$.

\subsection{Benefits of the Exemplar-based Framework}
\label{sec:exe}

Prior works on object recognition, e.g., \cite{malisiewicz2011ensemble}, adopt exemplar-models to infer the unknown objects. Similarly, our framework adopts exemplar-AMMs to evaluate the crowd movement. As discussed, the entropy value implies how well or poorly the AMM fits to the crowd data.
Because all the exemplar-AMMs contribute to the feature, the computed entropy values of exemplar-AMMs are jointly formed as a descriptor of the crowd movement.
The main advantage of this approach is to obviate the difficulty of training a robust AMM by leveraging the joint efforts of these pre-defined exemplar-AMMs.

Assuming that we have $K$ exemplar-AMMs, we then have an entropy matrix, $[\textbf{E}^1, ..., \textbf{E}^K] \in \mathbb{R}^{N \times K}$, where $\textbf{E}^i$ contains the entropies of the crowd individuals in reference to exemplar-AMM-$i$. In practice, we compute the average and the variance values of $\textbf{E}^k$ to form a feature vector, i.e. $[avg(\textbf{E}^1),std(\textbf{E}^1) ..., avg(\textbf{E}^K), std(\textbf{E}^K)] \in \mathbb{R}^{2K}$. Since we assume that the crowd members are different, the mean entropy value illustrates the average performance of the AMM in fitting to the query crowd data. The standard deviation of the entropy value measures how differently the AMM fits to the individual trajectories of the crowd. For example, the standard deviation should be small for a coherent motion, since the differences among crowd members are small, while it should be large for a random motion.

\section{Multi-label Classification}

As it is often difficult to assign a single label to any real world crowd movement, we treat the crowd movement recognition problem as a multi-label classification problem in this paper. Specifically, we let $\textbf{X}$ be the feature vector of an instance, e.g., a crowd movement sequence, and $\textbf{Y}$ be a finite set of labels, i.e. $\textbf{Y} = \{1,2,\dots,Q\}$. Given a training set $T = \{(x_1, Y_1), (x_2, Y_2), \dots, (x_m, Y_m)\}$ ($x_i \in \textbf{X}$, $Y_i \subseteq \textbf{Y}$), our goal is to output a multi-label classifier $h: \textbf{X} \to 2^{\textbf{Y}}$, which is usually formed as a real-value function $g: \textbf{X} \times \textbf{Y} \to \mathbb{R}$. The function $g(\cdot, \cdot)$ can be treated as a ranking function.
Hence, the multi-label classifier can be derived as $h = \{y \mid g(x_i, y) > threshold, y \in \textbf{Y}\}$.


\section{Experiments and Results}

In this section, we first present how we select exemplar-AMMs based on \textit{SynCrowd}. We then analyze the computed features of the simulation data. Finally, we show the results on the recognition of real-world crowd movement.


\begin{figure*}
		\centering
	\begin{tabular}{c@{}c@{}c}
		\includegraphics[width=.25\textwidth]{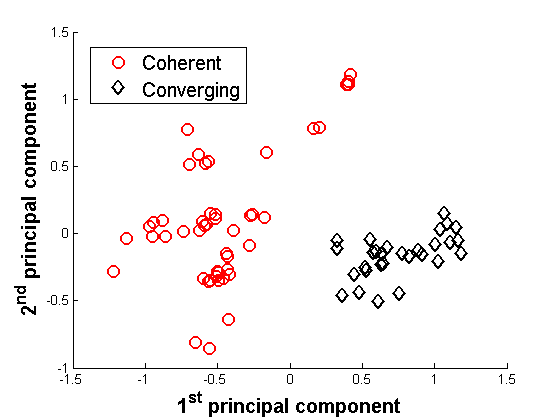} & \hspace{.3in}
		\includegraphics[width=.25\textwidth]{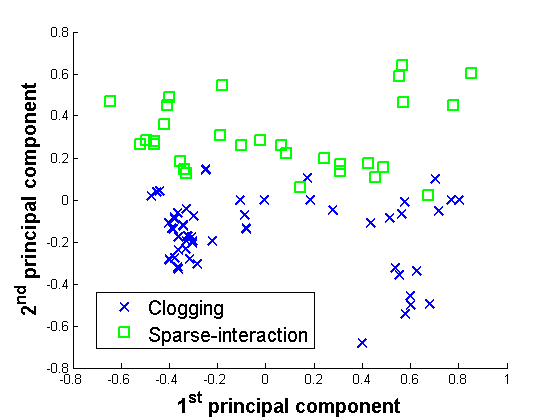} & \hspace{.3in}	
		\includegraphics[width=.25\textwidth]{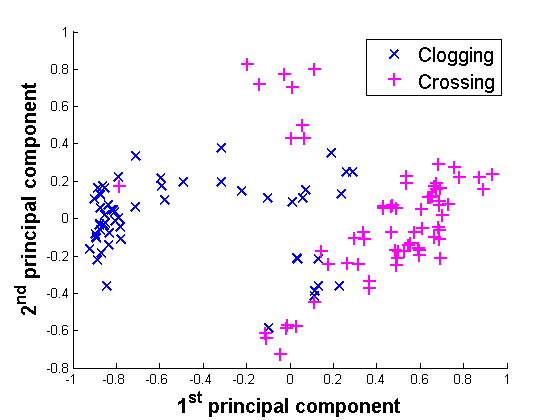} \\
		{\footnotesize (a) Coherent (coH) vs Converging (coL)} & \hspace{.3in} {\footnotesize (b) Clogging (cG) vs Sparse-interaction (sP)} & \hspace{.3in} {\footnotesize (c) Clogging (cG) vs Crossing (cR) } \\
	\end{tabular}
	\caption{Visualization of the features of different simulation data in the coordinate space of the top-2 principal components.}
	\label{fig:pca}
	\vspace{-.13in}
\end{figure*}

\subsection{Selection of Exemplar-AMMs}

To select exemplar-AMMs, we first set up a synthetic crowd motion dataset, called \textit{SynCrowd}.
We design and generate the crowd simulation data, including five categories of clustered motion: coherent motion, converging motion, sparse interaction, crossing, and clogging.
A coherent motion (coH) refers to the motion of individuals in the crowd generally moving in the same direction and maintaining almost the same distance from each other.
For the converging motion (coL), the crowd members move closer to each other.
The sparse interaction (sP) only exists in sparse crowds, where individuals move independently.
Crossing (cR) and clogging (cG) take place in a dense crowd with multiple groups.
We illustrate some examples in Fig.~\ref{fig:examples_motions}.


\begin{table}
	\caption{The key motion parameters of the selected AMMs.}
	\centering
	{
		\begin{tabular}{|c | c | c | c | c | c | c |}
			\hline
			& \textbf{\emph{N-Dist}} & \textbf{\emph{N-Num}} & \textbf{\emph{React}} &
            \textbf{\emph{Radius}} & \textbf{\emph{Max-Spd}} & \textbf{\emph{Smooth}} \\
			\hline\hline
			AMM-1 & 2.94 & 3 & 3.12 & 0.13 & 3.61 & 0.25 \\
			AMM-2 & 5.84 & 8 & 0.83 & 0.31 & 3.80 & 0.76 \\
			AMM-3 & 1.96 & 3 & 4.12 & 0.25 & 2.71 & 0.31 \\
			AMM-4 & 8.72 & 3 & 1.31 & 0.19 & 1.52 & 0.36 \\
			AMM-5 & 5.61 & 8 & 0.06 & 0.19 & 3.17 & 0.82 \\
			AMM-6 & 8.96 & 4 & 0.18 & 0.12 & 1.58 & 0.06 \\
			AMM-7 & 8.84 & 2 & 6.09 & 0.66 & 1.75 & 0.08\\
			\hline
		\end{tabular}
	}
	\label{tab:select_param}
\end{table}



Here, we adopt RVO~\cite{van_den_berg_reciprocal_2011} to form the AMMs. We take into consideration the following adjustable variables of RVO: \emph{N-Dist}, the neighborhood range; \emph{N-Num}, the number of nearest agents within a certain range; \emph{Radius}, the effective radius of an agent (i.e. the distance that the agent prefers to keep away from others); \emph{React}, the range that the agent travels to avoid an upcoming collision; \emph{Max-spd}, the maximum speed; and \emph{Smooth}, how smoothly the agent can change velocities. As we tweak these variables, different AMMs are created. We first need to generate several combinations of these 6 variables as candidates of the exemplar-AMMs.
By thresholding each variable, we can sample it from a reasonable range (e.g., the neighboring radius rests within [0.1, 2] meters). However, there are too many combinations if we evenly sample each variable within its valid range and combine them together.
RVO may sometimes not be sensitive to certain variables. For example, if \emph{N-Dist} is larger than 4 meters, its small variation may not make a big difference. Therefore, we empirically reduce the number of variable combinations down to 25 and then add random white noise to introduce randomness to them. 	
However, some of these candidate AMMs may still be redundant.
To eliminate the useless AMMs, we use the crowd data from \textit{SynCrowd} and compute the entropy descriptors using all candidate AMMs. We then apply the sequential feature selection algorithm~\cite{jain1997feature} and adopt the misclassification rate of SVM as the selection criterion to choose the AMMs that can separate the classes well. As a result, only 7 exemplar-AMMs are selected from the candidate AMMs and used in our experiments, as shown in Table~\ref{tab:select_param}.


\subsection{Evaluation Based on Simulation}
To evaluate the effectiveness of our descriptor, we first test it on simulated crowd data.

\textbf{PCA:} We perform principal component analysis on the simulation data. For different types of crowd data, we transform their selected AMM-based features to principal components. Each time, we compare two types of simulated crowd movements and visualize the top-2 principal components. Fig.~\ref{fig:pca} shows the representative ones.  We observe that most of the pairs are separable. However, there are a few points in Fig.~\ref{fig:pca}(c) that may not be separable, because our entropy descriptor may not work well in certain scenarios. Nevertheless, this experiment generally shows that our proposed feature can be used to classify crowd movements.

\begin{table}[]
	\centering
	\caption{Comparison of using different numbers of exemplar-AMMs.}
	\begin{tabular}{|c|c|c|c|c|}
		\hline
		& \textbf{Accuracy} & \textbf{Precision} & \textbf{Recall} & \textbf{F-measure}
           \\ \hline \hline
		1 AMM  & 0.461    & 0.327     & 0.416  & 0.340     \\
		2 AMMs & 0.617    & 0.543     & 0.575  & 0.528    \\
		3 AMMs & 0.685    & 0.651     & 0.672  & 0.633     \\
		4 AMMs & 0.729    & 0.704     & 0.712  & 0.681     \\
		5 AMMs & 0.800    & 0.790     & 0.787  & 0.766     \\
		6 AMMs & 0.881    & 0.875     & 0.882  & 0.864     \\
		7 AMMs & 0.924    & 0.928     & 0.934  & 0.923    \\
		\hline
	\end{tabular}
	\label{tab:amm_comp}
\end{table}

\textbf{Classification:} To further evaluate our method, we apply the SVM linear classifier to classify the simulation data in~\textit{SynCrowd} based on the 10-fold cross validation strategy.
Table~\ref{tab:amm_comp} shows the average accuracy, precision, recall, and F-measure in classifying the \textit{SynCrowd} data with respect to different numbers of AMMs used. We can see that all metrics of using 7 AMMs are over 90\% in categorizing these five interaction patterns, outperforming all others. We also observe that the more exemplar-AMMs that are used in the classification, the better the metrics it obtains.


\subsection{Evaluation on Real-world Crowd Motion}

Since our method is based on pedestrian trajectories, we adopt a multi-person tracker to capture the crowd trajectories from videos.
Specifically, we first manually provide the initial position of each pedestrian.
We then use the state-of-the-art tracker~\cite{STRUCK_CVPR_2010} to track pedestrians' positions to produce the trajectories. Occasionally, the tracker may not perform well due to problems such as occlusions. The tracker is manually reinitialized once the tracking deviation is too large.
Finally, the captured trajectories are transformed from the image-space to the ground-space based on the estimated perspective transformation matrix, as the input data for our algorithm.

We labeled crowd trajectories from 524 short crowd videos, selected and split from the dataset in~\cite{shao2014scene}, where the average crowd size is 19.4. There are two main differences between real-world crowd motion and the simulated \textit{SynCrowd} data. First, as real-world crowd movements are more complex, we assign more semantic labels for them, including coherent motion, loosely coherent motion, sparse interaction, dense interaction, converging, crossing, clogging, and random motion. Second, real-world crowd data may not be uniquely labeled. For example, the crowd movement in a crosswalk may be labeled as both `coherent motion' and `crossing'. Hence, we treat the labeling of real-world crowd data as a multi-label classification problem. Fig.~\ref{fig:examples_motions}(f) shows the scene-category statistics of the dataset.

In this experiment, we have adopted the ML-KNN classifier~\cite{Zhang:2007:MLL:1234417.1234635}, which utilizes the maximum a posteriori principle, to predict the labels of the real-world crowd movement.
As a baseline comparison, we adopt the MBH descriptor(implemented based on~\cite{uijlings2015video}) to classify the real-world crowd movement.
In addition, we also compare our method with the latest method~\cite{shao2014scene}, which computes a bundle of crowd movement descriptors based on the group motion and then combines the group descriptors as features for crowd movement classification. For fair comparison, we provide both KLT tracklets and complete crowd trajectories as input to their algorithm.
Besides, we use a genetic algorithm (GA)~\cite{wolinski2014parameter} to train an optimal RVO parameter for each instance and apply the learned parameters as the feature.
Further, we also compare our feature with the features computed by 1 AMM and 4 AMMs.
Finally, as mentioned in Sec.~\ref{sec:exe}, our feature vector consists of the average values and the standard deviation values. We separate the feature vector into two sub-vectors: one containing only the average values (7 AMMs(m)) and the other containing only the standard deviation values (7 AMMs(s)). We compare these two sub-vectors with the complete feature.



In these experiments, we adopt the leave-one-out cross validation strategy to evaluate the performance of the multi-label classification. We also follow the metrics introduced in \cite{Zhang:2007:MLL:1234417.1234635}: (1) \textit{average precision}; (2) \textit{Hamming loss}, i.e. how many times an instance-label pair is misclassified; (3) \textit{one-error}, i.e. how many top-ranked labels are not in the proper label sets; (4) \textit{coverage}, i.e. how many labels are needed to cover all the instances;
and (5) \textit{ranking loss}, i.e. how many label pairs are reversely ordered for the instance. Except for \textit{average precision}, the smaller the metric is, the better the performance.

Table~\ref{tab:results} shows the classification result.
First, the trajectory-based methods perform better than the MBH descriptor and KLT-based~\cite{shao2014scene}, as trajectories provide more information on crowd movement.
Second, our features outperform not only GA-trained features but also the features from \cite{shao2014scene}. \cite{shao2014scene} originally requires dense input data to capture the properties of the group motion. When dealing with trajectories, the group descriptors may not work particularly well as the input data is not that dense.
Besides, as discussed earlier, using multiple exemplar-AMMs helps boost the classification performance. This is because more exemplar-AMMs can improve the describability of the crowd movements. We have also noticed that 7 AMMs(s) performs better than 7 AMMs(m), which implies that the standard deviation of the entropy values plays an important role in classification.

\begin{figure*}
	\centering
	\hspace{-.1in}
	\begin{tabular}{c@{}c@{}c@{}c@{}c@{}c@{}c@{}c}
		\footnotesize{a)} &
		\includegraphics[width=.16\textwidth]{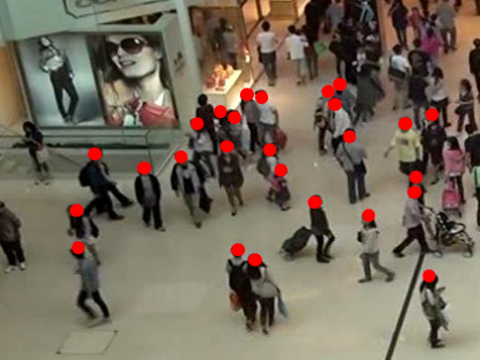} &	
		\includegraphics[width=.12\textwidth]{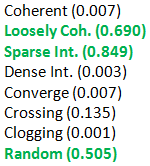}		&
		\includegraphics[width=.23\textwidth]{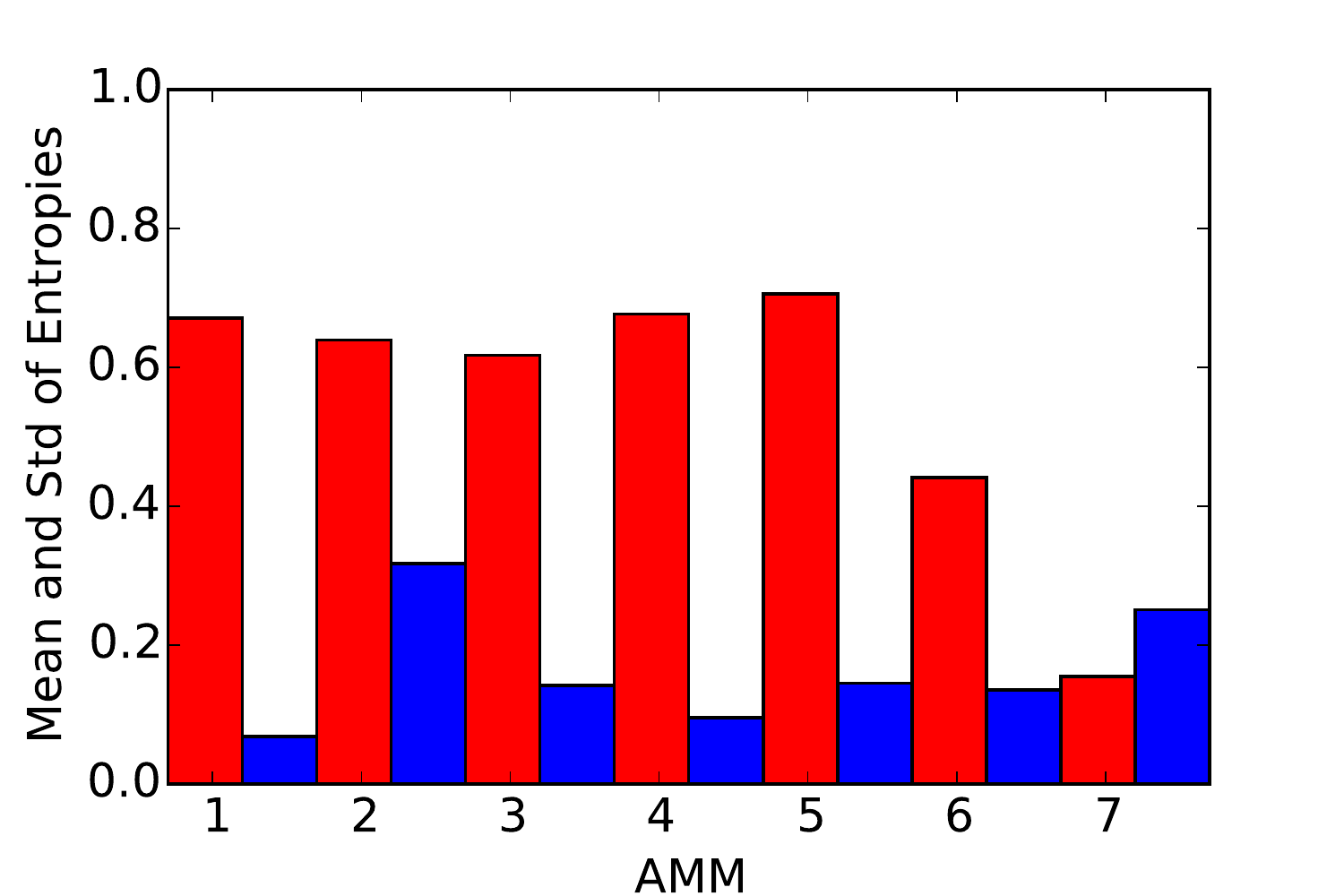}	&
		\footnotesize{b)} &
		\includegraphics[width=.16\textwidth]{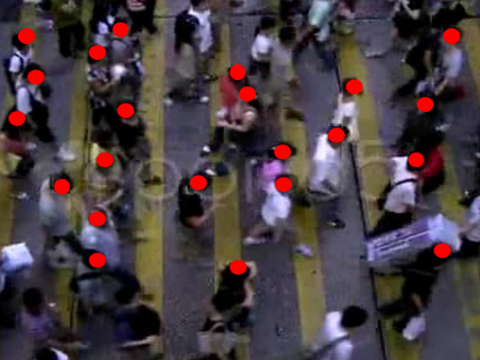}		&
		\includegraphics[width=.12\textwidth]{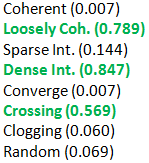} &
		\includegraphics[width=.23\textwidth]{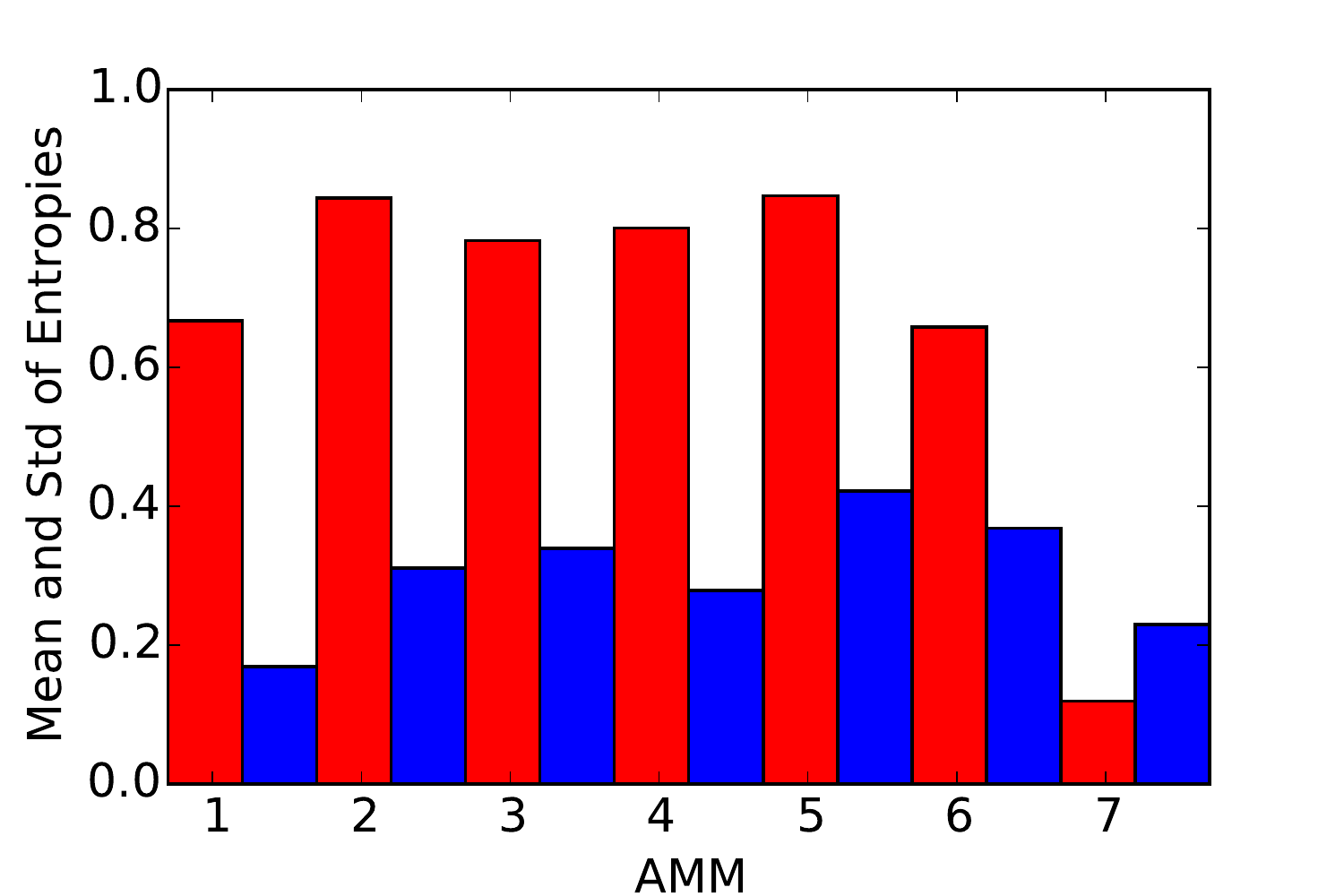}\\
		\footnotesize{c)} &
		\includegraphics[width=.16\textwidth]{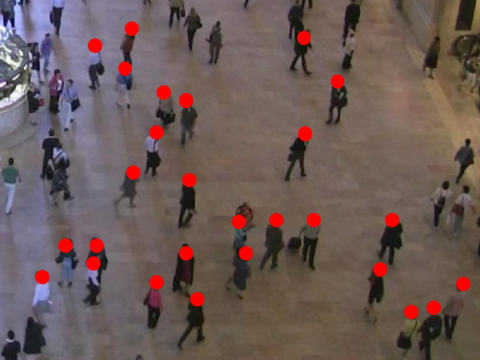} &
		\includegraphics[width=.12\textwidth]{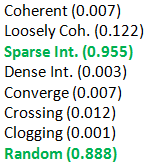}		&		
		\includegraphics[width=.23\textwidth]{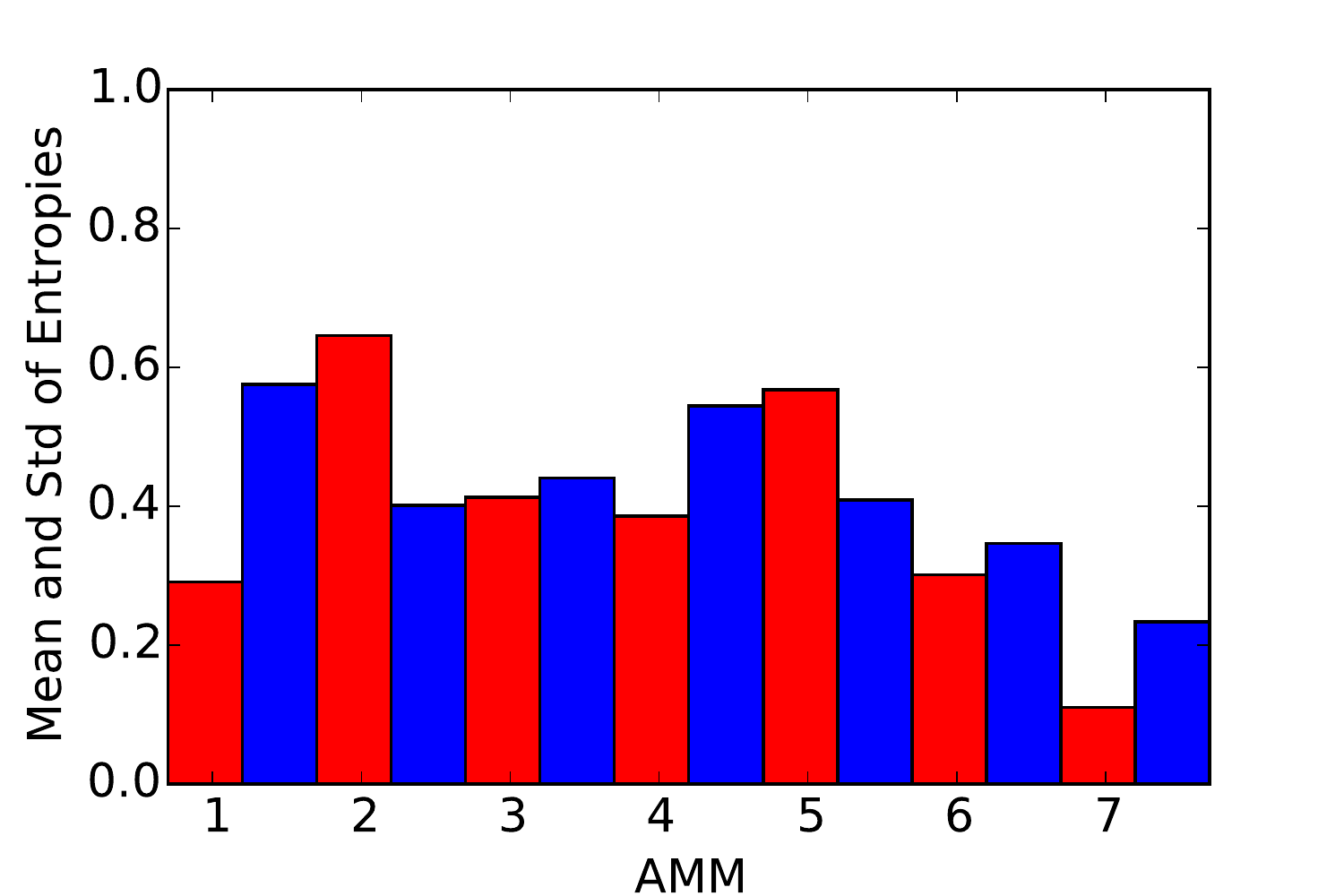}		&
		\footnotesize{d)} &
		\includegraphics[width=.16\textwidth]{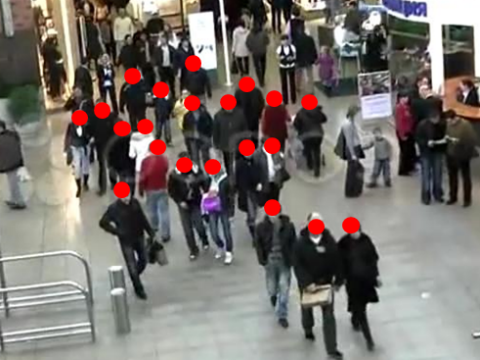}&
		\includegraphics[width=.12\textwidth]{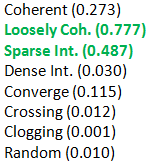}		&
		\includegraphics[width=.23\textwidth]{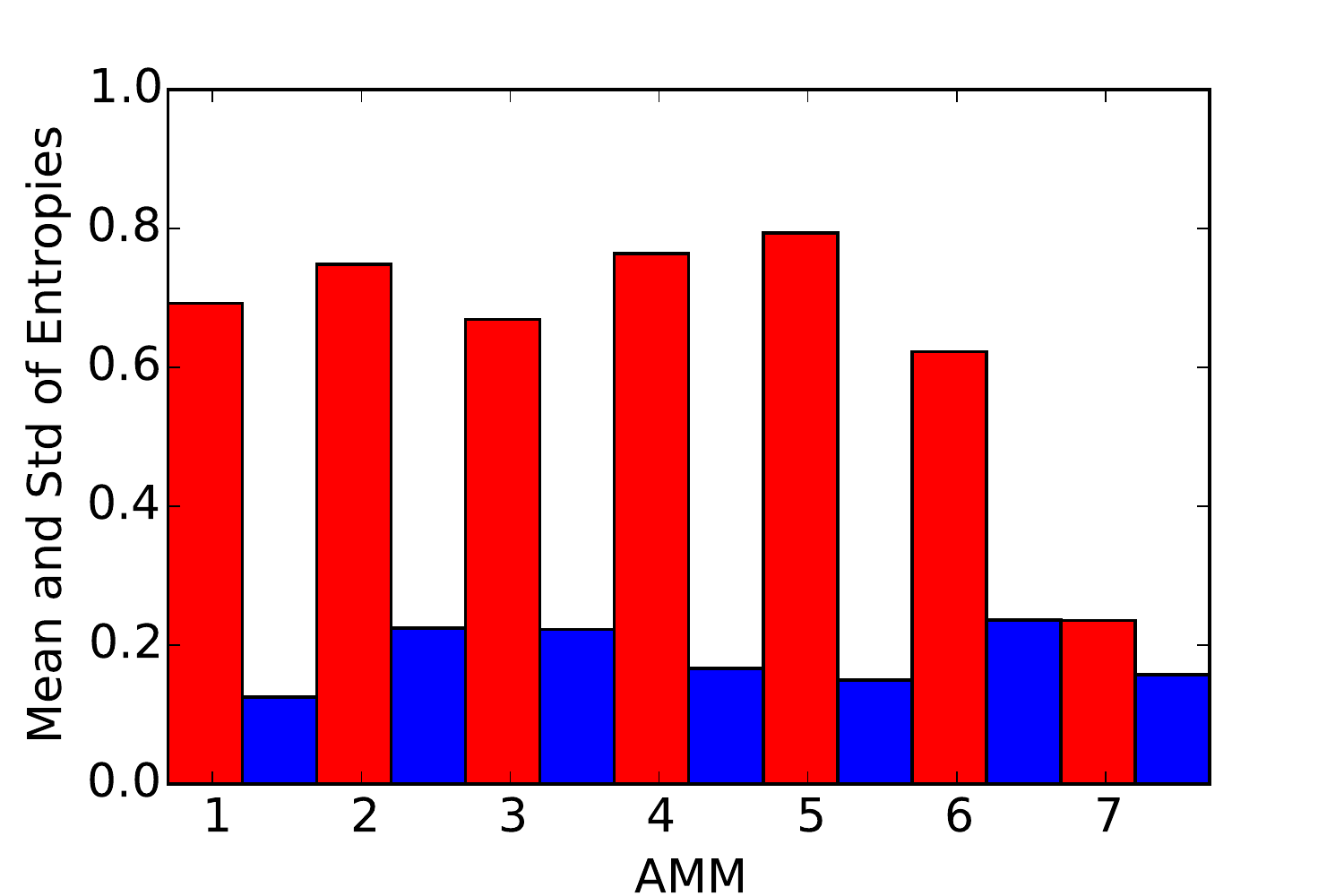}\\
		\footnotesize{e)} &
		\includegraphics[width=.16\textwidth]{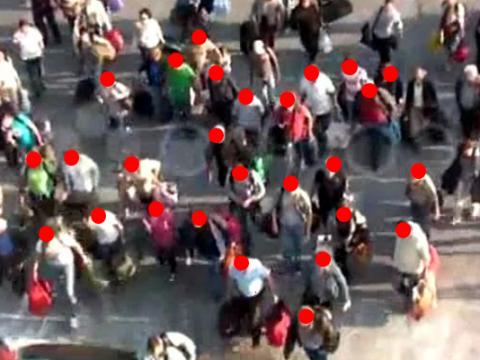} &
		\includegraphics[width=.12\textwidth]{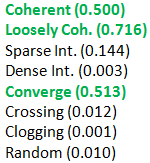}		&		
		\includegraphics[width=.23\textwidth]{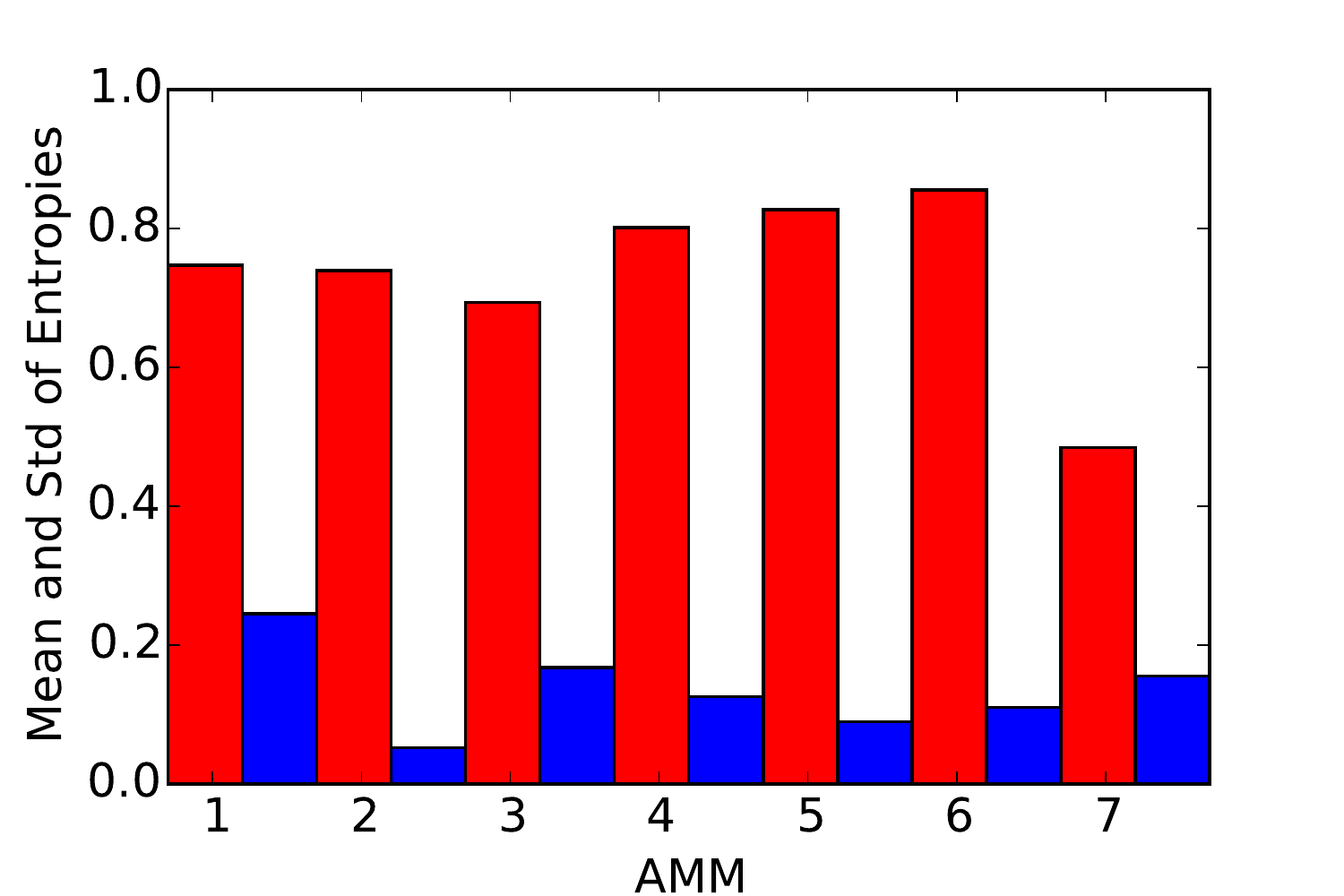} &
		\footnotesize{f)} &
		\includegraphics[width=.16\textwidth]{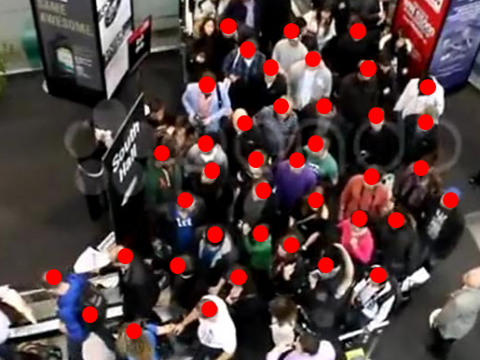}&
		\includegraphics[width=.12\textwidth]{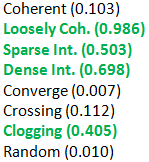}		&
		\includegraphics[width=.23\textwidth]{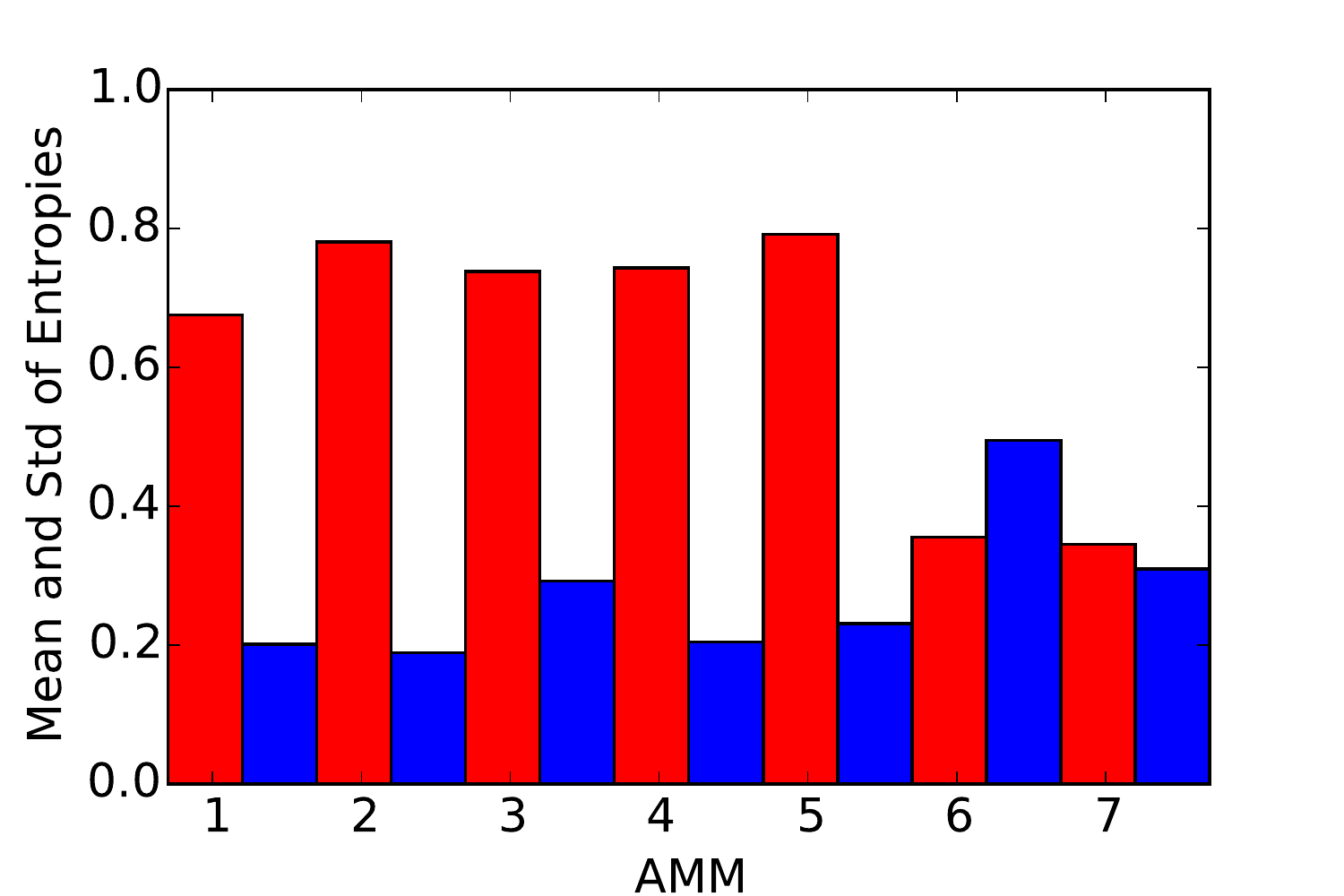} 		\\			
	\end{tabular}
	\caption{Results of classifying different real-world crowd videos using our feature. Red dots are the tracked pedestrians in the videos. The probabilities of estimating the crowd movement classes are also shown, and the bolded green ones are top-ranked classes. In addition, the normalized average (red bars) and the normalized standard deviation (blue bars) of the entropy values are computed via exemplar-AMMs in all the plots.
	}
	\vspace{-.2in}
	\label{fig:test}
\end{figure*}

We show more classification results on various crowd movements in Fig.~\ref{fig:test}. Specifically, Fig.~\ref{fig:test}(a)(c)(d) are captured from places like a mall and a train station, where mutual interactions of pedestrians take place at times (i.e. `sparse interaction'). The densities of Fig.~\ref{fig:test}(a) and (d) are higher than that of Fig.~\ref{fig:test}(c). This means that the motions of some pedestrians are more constrained and they are directed to move in the same direction (e.g., towards the exit). Hence, they are `loosely coherent.' Compared with Fig.~\ref{fig:test}(d), some people in Fig.~\ref{fig:test}(a) and most people in Fig.~\ref{fig:test}(c) are more heterogeneous. As a result, they are labeled as `random.' Note that the estimated probabilities of the labels are consistent with the scenarios. For example, Fig.~\ref{fig:test}(c) has a lower density than Fig.~\ref{fig:test}(a). As a result, its probability of being labeled `random' is much higher (0.888 $>$ 0.505) and its probability of being labeled `loosely coherent' is lower (0.122 $<$ 0.690).
In addition, Fig.~\ref{fig:test}(b)(e)(f) are correctly labeled as `crossing,' `converging,' and `clogging,' respectively, and their latent interaction and motion properties are consistent.

Finally, in Fig.~\ref{fig:test}, we also show the corresponding normalized average (red bars) and normalized standard deviation (blue bars) of the entropy values of AMMs.
According to Sec.~\ref{sec:exe}, a small average entropy value indicates that the AMM fits the crowd data well and vice versa, while a small standard deviation of the entropy values implies that the heterogeneity of the crowd movement is low. Results in Fig.~\ref{fig:test} are consistent with our expectation.
For example, Fig.~\ref{fig:test}(c) contains a low-density crowd, i.e., `random motion.' On the one hand, for a low-density crowd with few mutual collisions, AMMs generally perform better, i.e., the mean of their entropy values is lower than those of other examples. On the other hand, the motion in Fig.~\ref{fig:test}(c) is not very coherent. Hence, its standard deviation is much higher than that of the others. Another example is Fig.~\ref{fig:test}(e), which contains a converging motion. Most AMMs cannot fit well, but it can still be differentiated from others due to its large mean value and low standard deviation value. In addition, Fig.~\ref{fig:test}(b) and~\ref{fig:test}(d) have similar mean values, but the standard deviation values help differentiate if they are `crossing.'

\begin{table}[]
	\centering
	\caption{Comparison of multi-label classification.}
	\label{tab:results}
	\begin{tabular}{|c|c|c|c|c|c|}
		\hline
		& \textbf{Avg. Prec.}    & \textbf{Hamming}    & \textbf{One-error}    &
        \textbf{Cov.}     & \textbf{Ranking}    \\
		\hline\hline
		MBH~\cite{uijlings2015video} & 0.558 & 0.304 & 0.887 & 2.552 & 0.203\\
		Shao \emph{et al.}~\cite{shao2014scene} (KLT) & 0.719 & 0.214 & 0.379 & 2.528 & 0.162\\
		Shao \emph{et al.}~\cite{shao2014scene} (traj.) & 0.795 & 0.157 & 0.272 & 2.089 & 0.112 \\
		GA~\cite{wolinski2014parameter} & 0.790 & 0.142 & 0.292 & 2.071 & 0.112 \\
		1 AMM & 0.789 & 0.156 & 0.275 & 2.141 & 0.122\\
		2 AMMs & 0.825 & 0.147 & 0.200 & 1.857 & 0.087 \\
		4 AMMs & 0.902 & 0.082 & 0.136 & 1.403 & 0.040 \\
		7 AMMs(m) & 0.877 & 0.101 & 0.168 & 1.578 & 0.060\\		
		7 AMMs(s) & 0.897 & 0.088 & 0.134 & 1.462 & 0.049\\
		Ours (7 AMMs)  & \textbf{0.924} & \textbf{0.072} & \textbf{0.097} & \textbf{1.313} & \textbf{0.031} \\
		\hline
	\end{tabular}

\end{table}

\subsection{Implementation and Timing Performance}
Finally, we test the timing performance of the proposed method on a laptop with 8GB memory and a two-core 2GHz i7 CPU. The implementation was not optimized. Only RVO was compiled in C++ and the rest was built on Matlab R2013a. The complexity of our algorithm is $O(TMN)$, since the computation of $f$ is $O(N)$, where $N$ is the number of agents, and we need to calculate $M$ ensembles in EnKF and MLE.
In our experiments, we set the number of ensembles to 1,000 and the initial $\Sigma$ in \textbf{STEP-1} as a diagonal matrix with small diagonal entries (e.g., $diag[1e-3, \cdots, 1e-3]$). The algorithm requires around 30 loops to converge. The time cost per loop depends on the crowd size and the motion duration. In our experiments, the average time per loop is about 2.1 seconds.



\section{Conclusion and Future Works}

In this paper, we have proposed an exemplar-based method to extract features from the crowd trajectories and have shown that the proposed method outperforms the state-of-the-art methods in recognizing both simulated and real-world crowd movements.
For our future work, we would like to address the limitation in the method that requires trajectories in ground-space as input. Also, the entropy descriptor is not accurate when handling long crowd movements.

There are several other related problems to address in the future. First, our method can be extended to analyze trajectories from different sources by leveraging different simulators (e.g. vehicle simulators). Second, our algorithm is slow for real-time surveillance applications.
Finally, it would be interesting to integrate multi-target detection and tracking with crowd motion classification.

{\small
	\bibliographystyle{IEEEtran}
	\bibliography{all_refs,crowd_analysis}

\begin{thebibliography}{10}
\providecommand{\url}[1]{#1}
\csname url@samestyle\endcsname
\providecommand{\newblock}{\relax}
\providecommand{\bibinfo}[2]{#2}
\providecommand{\BIBentrySTDinterwordspacing}{\spaceskip=0pt\relax}
\providecommand{\BIBentryALTinterwordstretchfactor}{4}
\providecommand{\BIBentryALTinterwordspacing}{\spaceskip=\fontdimen2\font plus
\BIBentryALTinterwordstretchfactor\fontdimen3\font minus
  \fontdimen4\font\relax}
\providecommand{\BIBforeignlanguage}[2]{{%
\expandafter\ifx\csname l@#1\endcsname\relax
\typeout{** WARNING: IEEEtran.bst: No hyphenation pattern has been}%
\typeout{** loaded for the language `#1'. Using the pattern for}%
\typeout{** the default language instead.}%
\else
\language=\csname l@#1\endcsname
\fi
#2}}
\providecommand{\BIBdecl}{\relax}
\BIBdecl

\bibitem{Wu2014TMM}
J.~Wu and D.~Hu, ``Learning effective event models to recognize a large number
  of human actions,'' \emph{IEEE TMM}, 2014.

\bibitem{Yang2015TMM}
X.~Yang, T.~Zhang, C.~Xu, and M.~Hossain, ``Automatic visual concept learning
  for social event understanding,'' \emph{IEEE TMM}, 2015.

\bibitem{pellegrini_youll_2009}
S.~Pellegrini, A.~Ess, K.~Schindler, and L.~van Gool, ``You'll never walk
  alone: modeling social behavior for multi-target tracking,'' in \emph{Proc.
  ICCV}, 2009.

\bibitem{Yuan2015TMM}
Y.~Yuan, H.~Yang, Y.~Fang, and W.~Lin, ``Visual object tracking by structure
  complexity coefficients,'' \emph{IEEE TMM}, 2015.

\bibitem{Feris2012TMM}
R.~Feris, B.~Siddiquie, J.~Petterson, Y.~Zhai, A.~Datta, L.~Brown, and
  S.~Pankanti, ``Large-scale vehicle detection, indexing, and search in urban
  surveillance videos,'' \emph{IEEE TMM}, 2012.

\bibitem{Presti2012TMM}
L.~Presti, S.~Sclaroff, and M.~Cascia, ``Path modeling and retrieval in
  distributed video surveillance databases,'' \emph{IEEE TMM}, 2012.

\bibitem{ge2012group}
R.~Ge, Collins, ``Vision-based analysis of small groups in pedestrian crowds,''
  \emph{IEEE TPAMI}, 2012.

\bibitem{wang2011trajectory}
X.~Wang, K.~Ma, G.~Ng, and W.~Grimson, ``Trajectory analysis and semantic
  region modeling using nonparametric hierarchical bayesian models,''
  \emph{IJCV}, 2011.

\bibitem{zhou2012understanding}
B.~Zhou, X.~Wang, and X.~Tang, ``Understanding collective crowd behaviors:
  Learning a mixture model of dynamic pedestrian-agents,'' in \emph{Proc.
  CVPR}, 2012.

\bibitem{reynolds_flocks_1987}
C.~Reynolds, ``Flocks, herds and schools: A distributed behavioral model,'' in
  \emph{Proc. ACM SIGGRAPH}, 1987.

\bibitem{helbing_social_1995}
D.~Helbing and P.~Molnar, ``Social force model for pedestrian dynamics,''
  \emph{Physical Review E}, vol.~51, no.~5, 1995.

\bibitem{van_den_berg_reciprocal_2011}
J.~van~den Berg, S.~Guy, M.~Lin, and D.~Manocha, ``Reciprocal n-body collision
  avoidance,'' \emph{Robotics Research}, 2011.

\bibitem{guy2012statistical}
S.~Guy, J.~van~den Berg, W.~Liu, R.~Lau, M.~Lin, and D.~Manocha, ``A
  statistical similarity measure for aggregate crowd dynamics,'' \emph{ACM
  TOG}, 2012.

\bibitem{malisiewicz2011ensemble}
T.~Malisiewicz, A.~Gupta, and A.~Efros, ``Ensemble of exemplar-svms for object
  detection and beyond,'' in \emph{Proc. ICCV}, 2011.

\bibitem{chan2008modeling}
A.~Chan and N.~Vasconcelos, ``Modeling, clustering, and segmenting video with
  mixtures of dynamic textures,'' \emph{IEEE TPAMI}, 2008.

\bibitem{mahadevan2010anomaly}
V.~Mahadevan, W.~Li, V.~Bhalodia, and N.~Vasconcelos, ``Anomaly detection in
  crowded scenes,'' in \emph{Proc. CVPR}, 2010.

\bibitem{wang2009unsupervised}
X.~Wang, X.~Ma, and W.~Grimson, ``Unsupervised activity perception in crowded
  and complicated scenes using hierarchical bayesian models,'' \emph{IEEE
  TPAMI}, 2009.

\bibitem{kuettel2010s}
D.~Kuettel, M.~Breitenstein, L.~van Gool, and V.~Ferrari, ``What's going on?
  {Discovering} spatio-temporal dependencies in dynamic scenes,'' in
  \emph{Proc. CVPR}, 2010.

\bibitem{choi2012unified}
W.~Choi and S.~Savarese, ``A unified framework for multi-target tracking and
  collective activity recognition,'' in \emph{Proc. ECCV}, 2012.

\bibitem{li2013recognizing}
R.~Li, R.~Chellappa, and S.~K. Zhou, ``Recognizing interactive group activities
  using temporal interaction matrices and their {Riemannian} statistics,''
  \emph{IJCV}, 2013.

\bibitem{wolinski2014parameter}
D.~Wolinski, S.~Guy, A.~Olivier, M.~Lin, D.~Manocha, and J.~Pettr{\'e},
  ``Parameter estimation and comparative evaluation of crowd simulations,'' in
  \emph{Computer Graphics Forum}, vol.~33, no.~2, 2014, pp. 303--312.

\bibitem{charalambous2014data}
P.~Charalambous, I.~Karamouzas, S.~J. Guy, and Y.~Chrysanthou, ``A data-driven
  framework for visual crowd analysis,'' \emph{Computer Graphics Forum},
  vol.~33, no.~7, pp. 41--50, 2014.

\bibitem{ali_floor_2008}
S.~Ali and M.~Shah, ``Floor fields for tracking in high density crowd scenes,''
  in \emph{Proc. ECCV}, 2008.

\bibitem{ali2007lagrangian}
{S. Ali} and M.~Shah, ``A {Lagrangian} particle dynamics approach for crowd
  flow segmentation and stability analysis,'' in \emph{Proc. CVPR}, 2007.

\bibitem{mehran2009abnormal}
R.~Mehran, A.~Oyama, and M.~Shah, ``Abnormal crowd behavior detection using
  social force model,'' in \emph{Proc. CVPR}, 2009.

\bibitem{ali2013measuring}
S.~Ali, ``Measuring flow complexity in videos,'' in \emph{Proc. ICCV}, 2013.

\bibitem{zhou2013measuring}
B.~Zhou, X.~Tang, and X.~Wang, ``Measuring crowd collectiveness,'' in
  \emph{Proc. CVPR}, 2013.

\bibitem{shao2014scene}
J.~Shao, C.~Loy, and X.~Wang, ``Scene-independent group profiling in crowd,''
  in \emph{Proc. CVPR}, 2014.

\bibitem{reynolds_steering_1999}
C.~Reynolds, ``Steering behaviors for autonomous characters,'' in \emph{Proc.
  Game Developers Conference}, 1999.

\bibitem{Moussaid2011}
M.~Moussaid, D.~Helbing, and G.~Theraulaz, ``How simple rules determine
  pedestrian behavior and crowd disasters,'' \emph{Proceedings of The National
  Academy of Sciences}, vol. 108, 2011.

\bibitem{ondej_synthetic-vision_2010}
J.~Ondrej, J.~Pettre, A.~Olivier, and S.~Donikian, ``A synthetic-vision based
  steering approach for crowd simulation,'' \emph{{ACM} TOG}, 2010.

\bibitem{pettre_experiment-based_2009}
J.~Pettre, J.~Ondrej, A.~Olivier, A.~Cretual, and S.~Donikian,
  ``Experiment-based modeling, simulation and validation of interactions
  between virtual walkers,'' in \emph{Proc. SCA}, 2009.

\bibitem{Karamouzas2010}
I.~Karamouzas and M.~Overmars, ``Simulating human collision avoidance using a
  velocity-based approach,'' in \emph{Proc. VRIPYS}, 2010.

\bibitem{guy2011simulating}
S.~Guy, S.~Kim, M.~Lin, and D.~Manocha, ``Simulating heterogeneous crowd
  behaviors using personality trait theory,'' in \emph{Proc. SCA}, 2011.

\bibitem{lee2007group}
K.~Lee, M.~Choi, Q.~Hong, and J.~Lee, ``Group behavior from video: a
  data-driven approach to crowd simulation,'' in \emph{Proc. SCA}, 2007.

\bibitem{charalambous2014pag}
P.~Charalambous and Y.~Chrysanthou, ``The {PAG} crowd: A graph based approach
  for efficient data-driven crowd simulation,'' \emph{Computer Graphics Forum},
  vol.~33, no.~8, pp. 95--108, 2014.

\bibitem{bera2015gi}
A.~Bera, S.~Kim, and D.~Manocha, ``Efficient trajectory extraction and
  parameter learning for data-driven crowd simulation,'' in \emph{Proc.
  Graphics Interface}, 2015.

\bibitem{yamaguchi_who_2011}
K.~Yamaguchi, A.~Berg, L.~Ortiz, and T.~Berg, ``Who are you with and where are
  you going?'' in \emph{Proc. CVPR}, 2011.

\bibitem{wenxiliu2015tracking}
W.~Liu, A.~Chan, R.~Lau, and D.~Manocha, ``Leveraging long-term predictions and
  online learning in agent-based multiple person tracking,'' \emph{IEEE TCSVT},
  2015.

\bibitem{antonini_behavioral_2006}
G.~Antonini, S.~Martinez, M.~Bierlaire, and J.~Thiran, ``Behavioral priors for
  detection and tracking of pedestrians in video sequences,'' \emph{IJCV},
  2006.

\bibitem{vicsek1995novel}
T.~Vicsek, A.~Czir{\'o}k, E.~Ben-Jacob, I.~Cohen, and O.~Shochet, ``Novel type
  of phase transition in a system of self-driven particles,'' \emph{Physical
  Review Letters}, 1995.

\bibitem{evensen2003ensemble}
G.~Evensen, ``The ensemble {Kalman} filter: Theoretical formulation and
  practical implementation,'' \emph{Ocean Dynamics}, vol.~53, no.~4, 2003.

\bibitem{jain1997feature}
A.~Jain and D.~Zongker, ``Feature selection: Evaluation, application, and small
  sample performance,'' \emph{IEEE TPAMI}, 1997.

\bibitem{STRUCK_CVPR_2010}
S.~Hare, A.~Saffari, and P.~Torr, ``Struck: Structured output tracking with
  kernels,'' in \emph{Proc. ICCV}, 2011.

\bibitem{Zhang:2007:MLL:1234417.1234635}
M.~Zhang and Z.~Zhou, ``{ML-KNN}: A lazy learning approach to multi-label
  learning,'' \emph{Pattern Recognition}, 2007.

\bibitem{uijlings2015video}
J.~Uijlings, I.~Duta, E.~Sangineto, and N.~Sebe, ``Video classification with
  densely extracted {HOG/HOF/MBH} features: an evaluation of the
  accuracy/computational efficiency trade-off,'' \emph{IJMIR}, 2015.

\end{thebibliography}
}

%
%

%

%
%
%




\end{document}